\documentclass[10pt, a4paper]{article}
\usepackage{lrec2022} 
\usepackage{multibib}
\newcites{languageresource}{Language Resources}
\usepackage{graphicx}
\usepackage{tabularx}
\usepackage{soul}
\usepackage{xspace}
\usepackage{titlesec}
\titleformat{\section}{\normalfont\large\bfseries\center}{\thesection.}{1em}{}
\titleformat{\subsection}{\normalfont\SmallTitleFont\bfseries\raggedright}{\thesubsection.}{1em}{}
\titleformat{\subsubsection}{\normalfont\normalsize\bfseries\raggedright}{\thesubsubsection.}{1em}{}
\renewcommand\thesection{\arabic{section}}
\renewcommand\thesubsection{\thesection.\arabic{subsection}}
\renewcommand\thesubsubsection{\thesubsection.\arabic{subsubsection}}

\usepackage{epstopdf}
\usepackage[utf8]{inputenc}

\usepackage{hyperref}
\usepackage{xstring}
\usepackage{float} 
\usepackage{color}
\usepackage{arabtex}
\usepackage{utf8}
\usepackage{makecell}
\usepackage{multirow}

\usepackage{tipa}
        \usepackage{arabtex}
        \usepackage{utf8}
        \usepackage{tipa}

\newcommand{\Ar}[1]{{\scriptsize \<#1>\xspace}}

\newcommand{\TrNoAr}[1]{\arabfalse\transtrue \RL{\Ar{#1}}\arabtrue\transfalse}

\newcommand{\TrAr}[1]{\arabtrue\transfalse {\scriptsize \Ar{#1}} \arabfalse\transtrue \RL{#1}\arabtrue\transfalse } 

\newcommand{\TrArSm}[1]{\Ar{#1}}

\settrans{english}

\def\correspondingauthor{\thanks{* Corresponding author.}}

\title{ {\cal Curras + Baladi}: Towards a Levantine Corpus}

\name{Karim El Haff, Mustafa Jarrar\correspondingauthor{*}, Tymaa Hammouda, Fadi Zaraket} 

\address{University of Strasbourg, Birzeit University, American University of Beirut \\
         karim.el-haff@etu.unistra.fr, mjarrar@birzeit.edu, 1171779@student.birzeit.edu, fz11@aub.edu.lb\\
         }

\abstract{
This paper presents two-fold contributions: a full revision of the Palestinian morphologically annotated corpus (Curras), and a newly annotated Lebanese corpus (Baladi). Both corpora can be used as a more general Levantine corpus. Baladi consists of around 9.6K morphologically annotated tokens. Each token was manually annotated with several morphological features and using LDC’s SAMA lemmas and tags. The inter-annotator evaluation on most features illustrates 78.5\% Kappa and 90.1\% F1-Score. Curras was revised by refining all annotations for accuracy, normalization and unification of POS tags, and linking with SAMA lemmas. This revision was also important to ensure that both corpora are compatible and can help to bridge the nuanced linguistic gaps that exist between the two highly mutually intelligible dialects. Both corpora are publicly available through a web portal.
 \\ \newline }

\begin{document}
\setcode{utf8}

\maketitleabstract
\section{Introduction}
The processing of the Arabic language is a complex field of research. This is due to many factors, including the complex and rich morphology of Arabic, its high degree of ambiguity, and the presence of several regional varieties that need to be processed while taking into account their unique characteristics. When its dialects are taken into account, this language pushes the limits of NLP to find solutions to problems posed by its inherent nature. It is a diglossic language; the standard language is used in formal settings and in education and is quite different from the vernacular languages spoken in the different regions and influenced by older languages that were historically spoken in those regions. 
Indeed, Arabic speakers use those local varieties in day-to-day communication. 
We can distinguish several families of dialects: Moroccan, Egyptian, Sudanese, Levantine, Iraqi and Khaliji (Gulf). Arabic dialects tend to diverge from Modern Standard Arabic (MSA) in terms of phonetics, morphology, syntax and vocabulary. 

Arabic content was mainly written in MSA. Recently, dialectal content has been increasing  massively, especially on social media. MSA is considered among the under-resourced languages by the NLP community \cite{DH21}. Dialectal Arabic (DA) is even less resourced. The resource gap between MSA and the dialects implies a large margin of error when MSA tools are used against dialectal content~\cite{zbib-etal-2012-machine}. Thus, it is important to build resources and tools to identify dialects in context and to treat Arabic content based on its {\em unique} dialectal identity. 

In this research, we focus on the Lebanese variety of Levantine Arabic, which is used in daily conversations and in the Lebanese media. It is spoken by about 6 million locals, and almost double that number in diaspora. The paper presents a morphologically annotated corpus for Lebanese. The development of the corpus uses texts covering a wide spectrum of subjects and registers. The corpus is designed to be compatible with, and leverage, Curras~\cite{JHRAZ17}, the Palestinian corpus with morphological annotations. In this way, both corpora can be used as a more general Levantine corpus, especially that the Palestinian dialect represents Southern Levantine and that Lebanese represents Northern Levantine varieties. In addition to providing new Lebanese corpus annotations, we have also revised Curras annotations to ensure compatibility with the LDC’s SAMA tags and lemmas ~\cite{MaamouriSama2010}.



In this paper, we present two-fold contributions: 
\begin{enumerate}
\item Baladi, a Lebanese morphologically annotated corpus, which consists of 9.6K tokens. Each token was manually annotated with prefixes, suffixes, stem, POS tags, MSA and DA lemmatization, English gloss, in addition to other features such as gender, number, aspect, and person. The corpus was annotated mainly using LDC’s SAMA lemmas and tags. 
The inter-annotator evaluation on most features illustrates 87\% agreement using the Cohen's Kappa score~\cite{j:kappa:McHugh2015}. 

\item Revision of Curras, by refining all annotations for accuracy,  normalization and unification of POS tags, and linking with SAMA lemmas. This revision was also important to ensure that both corpora are compatible and can together form a more general Levantine corpus. 
\end{enumerate}

Both, Curras and Baladi, are publicly available online\footnote{\scriptsize{\url{https://portal.sina.birzeit.edu/curras}}}.\\

The rest of the paper is organized as follows. We overview related work in Section \ref{sec:related_work}. Section \ref{sec:lebanese} describes the Lebanese dialect. In Section \ref{sec:corpus_collection}, we present Corpus Baladi. In Section \ref{sec:annotation_methodology} we present the annotation process and guidelines. Section \ref{sec:eval} presents the evaluation and the inter-annotator agreement. In Section \ref{sec:Curras_Revisions}, we present the revisions we introduce to Curras. Section \ref{sec:discussion} discusses how we managed to transform Curras into a  more Levantine corpus. Finally, Section \ref{sec:conclusion} concludes.

\section{Related Work}
\label{sec:related_work}
This section reviews efforts to create annotated corpora for Arabic dialects as well as for MSA.

\subsection{MSA Resources} 
The Penn Arabic Treebank (PATB)~\cite{MaamouriPATB2005} by the Linguistic Data Consortium (LDC) is central to the development of several MSA resources. It enriches newswire text in MSA collected from several news outlets with tokenization, segmentation, lemma, POS and gloss tags annotations along with syntactic trees. PATB uses the morphological tags as defined by the BAMA morphological analyzer~\cite{BuckwalterBAMA2004}, which provides vocalized solutions, unique lemmas, prefixes, suffixes, stems, POS tags, and English gloss terms. SAMA~\cite{MaamouriSama2010} is a substantial improvement and refinement on BAMA as it extends its lexicon and provides several analysis refinements. 
The Prague Arabic Dependency Tree bank~\cite{PragueADT_2004} enriched the literature with functional linguistic annotations which in turn lead to the emergence of ElixirFM~\cite{smrz-2007-elixirfm} 
The Arabic lexicographic database at Birzeit University \cite{JA19,ADJ19} provides a large set of MSA lemmas, word forms, and morphological features, which are linked with the Arabic Ontology \cite{J21,J11} using the W3C LEMON model \cite{JAM19}

\subsection{Dialectal Resources} 
The Levantine Arabic Treebank~\cite{maamouri-etal-2006-developing} featured the Jordanian Arabic dialect. Curras~\cite{JHRAZ17,JHAZ14} is a more recent Levantine corpus featuring the Palestinian dialect. Large number of textual entries were collected from Facebook, Twitter and scripts of the Palestinian series ``Watan Aa Watar''.Each word in the corpus was then manually annotated with a set of morphological attributes. The corpus contains  56K tokens. 

Earlier, the CALLHOME Egyptian Arabic corpus ~\cite{CallHome_1997} consisted of transcripts of telephone conversations in Egyptian. CALIMA~\cite{habash-etal-2012-morphological} extended ECAL~\cite{EcalKilany_2002} which build on CALLHOME to provide morphological analysis functionality of the Egyptian dialect. The COLABA project~\cite{Colaba_diab_2010} 
collected resources in dialectal Arabic (mainly in Egyptian and Levantine) from the collection of online blogs. The effort eventually lead to constructing the Egyptian Tree Bank (ARZATB) \cite{maamouri-etal-2014-developing}. Curras and ARZATB were leveraged as case studies for morphological analysis and disambiguation~\cite{eskander-etal-2016-creating}. 
YADAC~\cite{al-sabbagh-girju-2012-yadac} focuses also on the Egyptian dialect identification and provides a mutli-genre approach. It is a collection of web blogs, micro blogs, and several Egyptian content discussion forums. 
MADAR~\cite{bouamor-etal-2014-multidialectal} is an ongoing multi-dialect corpora covering 26 different cities and their corresponding dialects. Other efforts cover Emirati~\cite{Ntelitheos2017,khalifa-etal-2018-morphologically}, Tunisian and Algerian~\cite{ZribiEBB15,Harrat2014BuildingRF},  
and Yemeni and Moroccan~\cite{al-shargi-etal-2016-morphologically}. 

Our proposed contributions in this paper is to enrich Curras by (1) providing a Lebanese Levantine extension and by (2) refining and revising Curras entries to better accommodate the general Levantine dialect.

\section{Lebanese and Levantine Dialects}
\label{sec:lebanese}

The Levantine family of dialects can be linguistically split into Northern Levantine including the Lebanese and Syrian dialects, and Southern Levantine including Palestinian and Jordanian. During the spread of Arabic from the seventh century onwards, the Levant was a region that mainly spoke Western Aramaic~\cite{skaf:tel-01368247}. Aramaic is a Semitic language continuum spoken mainly during antiquity throughout the Levantine region and it served as a lingua franca then. Aramaic survives today through modern dialects such as Turoyo Syriac and Western Neo-Aramaic spoken in parts of Syria. It also survives more subtly in the noticeable substratum underlying Levantine dialects that differ from MSA on several common linguistic specifities such as phonology, syntax, morphology and lexicon. This motivates using dialect specific annotations to annotate Levantine dialects. In the sequel, we briefly review the differentiating factor between Levantine dialects and MSA. 

\subsection{Phonological differences} 
Like other Semitic languages, Aramaic and its varieties were written with a 22-letter alphabet (Abjad). When Arabic was spread to the Levant, the Christian populations of the region began to transcribe the Arabic language using this consonantal alphabet, a tradition of Syriac writing known as "Garshouni"~\cite{briquelchatonnet:hal-00278273}. Due to the lack of some letters compared to the Arabic alphabet which contains 28 letters, adaptations were made in the Garshouni script, and some Syriac graphemes can represent several phonemes of Arabic, especially among the emphatic letters. Indeed, certain Arabic phonemes were not widely used by Levantine populations, even to this day; a speaker of Lebanese today tends to de-emphasize emphatic letters in Arabic words as for example in the case of \Ar{مظلوم} (abused) which is pronounced  \TrNoAr{مظلوم} in MSA and \TrNoAr{مزلوم} in Lebanese. Another example may be words containing \TrAr{ث} such as \Ar{ثعلب} (fox) that is pronounced \TrNoAr{تعلب} or \TrNoAr{سعلب} across the Levant. As phonology differs in many situations for Levantine dialect speakers, spelling can vary greatly and can pose a challenge to the processing of those dialects when written.

\subsection{Syntactical differences} 
A common usage for sentence structure in MSA is the verb-subject-object (VSO) structure. Sentences tend to start with a verb followed by its subject and then its object. Other structure configurations tend to be less frequent. On the other hand, in Levantine dialects, this structure is more flexible as the verb and subject have a natural flow of interchangeable positions.
\begin{tabular}{lr}
MSA (VSO):& \Ar{أكل الولد التفاحة }\\
LEVANTINE (VSO):& \Ar{أكل الولد التفاحة} \\
LEVANTINE (SVO):& \Ar{ الولد أكل التفاحة } \\
In English (SVO):& The child ate the apple
\end{tabular} 

\subsection{Morphological differences} 
Levantine inherits templatic morphology where affixes play an important role from its Semitic roots. Major morphological differences exist when compared to MSA. One of them is the loss of case markings in Levantine. Additionally, there are Levantine-specific morphemes that do not exist in MSA such as \TrAr{عم};  the present continuous mark that precedes imperfect verbs to indicate the continuity of the action. Its absence in Levantine indicates that the action is a general truth: \Ar{أنا عم باكل } (I am eating) \Ar{أنا باكل} (I eat).  MSA has no such entity. Context alone indicates whether the action is continuous or a general truth; \Ar{أنا آكل} can mean both  ``I am eating at the moment'' or ``I factually eat''. Other morphemes include \TrAr{رح} and \TrAr{ح} that are the future markers in Levantine dialects as opposed to MSA’s \TrAr{س} and \TrAr{سوف}. 
Furthermore, the progressive Levantine particle \TrAr{ب} (as in \TrAr{باكل}) that is used to indicate imperfective verbs does not exist in MSA.

\subsection{Differences in lexicon} 
It is also important to notice that the Levantine lexicon is rich with old Aramaic words due to its pre-Arab heritage, as well as foreign loan words due to the Levant's location as a frequent passage of many civilisations.

\section{Corpus Collection}
\label{sec:corpus_collection}

We manually collected texts written in Lebanese from sources such as Facebook posts, blog posts and traditional poems. 
We collected a total of 9.6K tokens spanning over 424 sentences. We merged them all into a single text file, with an average of 22 words per sentence. The corpus was chosen based on a critical judgment to include several registers of Lebanese speech, hence the choice to include folk poems, and satirical texts from social networks and blog articles. We avoided text written in Arabizi (Arabic written using proprietary Latin letters) as this is not the goal of our corpus at this phase. As the size of the corpus is relatively small, we performed data collection manually through the retrieval of the transcripts of traditional Lebanese poems \TrAr{زجل} by local poet Jihad Assi, satirical Facebook posts written in the vernacular Lebanese dialect by Mohamad Jaber as well as some blog posts written in the Lebanese dialect (Bel-Lebneene blog).

We did not preprocess the text and kept the raw form. As such, we did not perform any unification of letter variations, removal of diacritics, or correction of typos. We based this on the selective quality of our corpora. We then tokenized the raw text. This produced a table with three columns (sentence ID, token ID, and token text). The table was represented in a modern shareable spread sheet tool where each token and its annotations stood on its own separate row. The annotators introduced annotations in separate columns each designated for a specific feature or tag.

\section{Annotation Methodology}
\label{sec:annotation_methodology}

\begin{table*}[ht]
    \centering
    \includegraphics [scale=0.9]{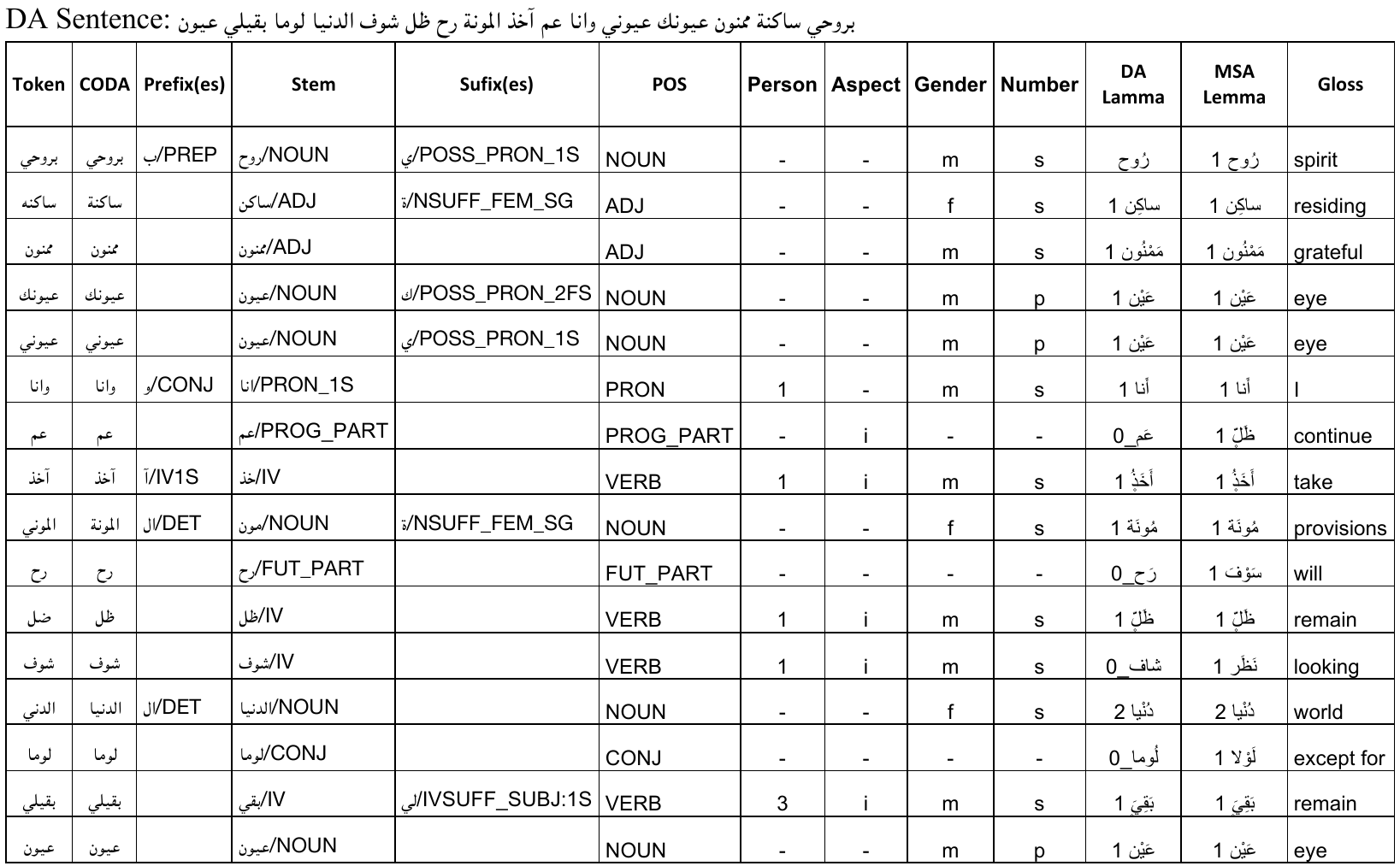}
    \caption{A sentence in Lebanese with its full annotations. Translation: ``My eyes are grateful to yours who dwell in my soul as I make my provisions and I shall still, even without my eyes, see the world whole.''}
    \label{table:Leb_sentence}
\end{table*}

Four linguists carried out the annotation process over a period of ten months. We used AnnoSheet to carry out the manual annotations. AnnoSheet is a Google Sheet that we empowered by developing and adding advanced JavaScript methods to (1) assist the annotation process and (2) validate the proposed annotations. 
For each row in the AnnoSheet, we have 16 columns: {\em sentence ID, Token ID, token, CODA, prefix, stem, suffix, POS, MSA lemma, dialect lemma, gloss, person, aspect, gender, and number}. The annotation guidelines for each of these columns are described in the following subsections. 

To speed up the annotation process, we uploaded the revised version of Curras annotations (See section \ref{sec:Curras_Revisions}) into another spreadsheet and allowed the annotator to look up candidate annotations from Curras. The JavaScript lookup method searches Curras and returns the top matching results. The annotator can then select one of the results, and edit the corresponding fields if needed. The annotator also has the option to fill in the annotations directly. To guide and control the quality of the annotations, we implemented several validation triggers in JavaScript to highlight potential mistakes. In addition, for each cell in AnnoSheet, we implemented a customized list, from which the annotator can select values based on the column. The lists are dynamic and they are populated with values that depend on the values in other cells of the same row.

We, additionally, developed a Google Colab application to validate all annotations in the AnnoSheet and to flag cells that may require corrections. The validation ran once daily. In this way, the annotators were able to annotate each word, in context, by re-using annotations from Curras, fully or partially, or by entering new annotations. As a Google Sheet, AnnoSheet allowed the annotators to also annotate the corpus cooperatively, write feedback to each other, and skip tokens they are not certain about. Table~\ref{table:Leb_sentence} illustrates a sentence in Lebanese and its full annotations.

\subsection{Annotation Guidelines } 
This section presents the annotation guidelines for each of the different annotations tasks.

\subsubsection*{CODA}
The CODA tag (\Ar{التهجئة الاصطلاحية}) of a token signifies the ``correct'' spelling of the token. Instead of annotating the exact token in the corpus, the idea is to unify the different spelling variations of the same word into one CODA spelling, and annotate this CODA. Due to the lack of standardized orthographic spelling rules for Arabic dialects, people tend to write words as they pronounce them; thus, the same word might be
written in different ways. In fact, the same person may write the same word in different ways in the same sentence. For example, consider the word meaning 'a lot': which can be written as \TrAr{كثير} and \TrAr{كتير}. The second letter in the first word corresponds to the sound $\theta$. The correct spelling in MSA is the first word with \Ar{ث}. However, the \Ar{ث} [th] sound is rarely pronounced in Lebanese and is often replaced by the \Ar{t} [t] sound or the [s] sound. Similarly, the words \TrAr{بألكن} and \TrAr{بقلكون} are different orthographic variations of the same word, which means ``I tell you''. In this case, we write both in the the same CODA spelling \TrAr{بقلكن}. See more examples in Table~\ref{t:coda_examples}. 
\begin{table}[H]
\centering
\begin{tabular}{ccc}
\thead{Token} &  & \thead{CODA} \\  \hline
 \TrAr{بألكن} &  → & \TrAr{بقلكن} \\
\TrAr{بقلكون} &  → & \TrAr{بقلكن}  \\
\TrAr{طريء} &  → &  \TrAr{طريق} \\
\TrAr{هايدي} &  → &  \TrAr{هيدي} \\
\TrAr{عيونن} &  → & \TrAr{عيونٌ}
\end{tabular} 
\caption{Example of words and their CODA spelling.} 
\label{t:coda_examples} 
\end{table} 

Since our goal in this corpus is to transform Curras to be a more general Levantine corpus, we chose to adopt the Palestinian CODA guidelines \cite{HJ15} for the Lebanese dialect. We made some modifications and simplifications to be adapted to cover more Levantine regionalisms, as will be discussed in the next sections. 

It is notable to add that some slight spelling differences exist between Lebanese and Palestinian as the former is a northern Levantine variety while the latter is  southern and regional differences exist. The most common examples of this lie in demonstrative pronouns where Palestinian tends to use more emphatic sounds than Lebanese; a masculine ``this'' is said \TrAr{هاظا} or \TrAr{هادا} in Palestinian, while \TrAr{هَيدا} in Lebanese. Another example is the use of \TrAr{م} in Palestinian to indicate a third person plural where  Lebanese uses a \TrAr{ن}: ``your house'' is \TrAr{بيتكم} in Palestinian and \TrAr{بيتكن} in Lebanese. The differences in spelling are due to the differences in pronunciation across the Levant and have no effect over the total mutual intelligibility of the dialects and thus a potential standardized spelling for Northern and Southern Levantine can be seen as the slight differences between British English and American English spelling systems.   

\subsubsection*{MSA Lemma}
The MSA Lemma (\Ar{المدخلة المعجمية الفصحى}) is the MSA lemma of the token. We restricted the choices of MSA lemmas to SAMA lemmas. The AnnoSheet allows the annotator to search the SAMA database and select the target lemma. For tokens that are not derived from an MSA lemma, like \TrAr{بدي}, we chose the closest SAMA lemma (e.g., 1\_\Ar{أَراد}).
In case, no matching MSA lemmas are found in the SAMA database, the annotator is allowed to look up lemmas from Birzeit's lexicographic database \cite{JA19}, which are linked with the Arabic Ontology \cite{J21} and represented in the w3C Lemon model \cite{JAM19}. the annotator may also introduce a new MSA lemma, however, new lemmas are marked with ``\_0'', such as (0\_\Ar{يوغا}) or (0\_\Ar{هيفاء}). Similar to SAMA lemmas, noun lemmas should be in the masculine singular form. Plural and feminine are acceptable in case there is no masculine singular. Verb lemmas should be in the past masculine singular 3rd person form. 

\subsubsection*{Dialect Lemma}
The dialectal lemma (\Ar{المدخلة المعجمية العامية}) signifies the semantic value of the token as a lexicon entry. Similar to the MSA lemma, each token in the corpus is tagged with its DA lemma. If a token stems from MSA, then its MSA and DA lemmas are the same. For example, the dialect token \TrAr{بقلك}, which means ``I tell you'', has the same MSA and DA lemma 1\_\Ar{قال}. Some Levantine lexicon instances differ from MSA and need their own dialectal lemmas. These lemmas potentially do not exist in an ordinary MSA dictionary, due to their likely origin in other languages, notably Aramaic. As an example, the typical Levantine words used to say 'inside' and 'outside' are \TrAr{جوّا} and \TrAr{برّا}, respectively. These two words are different in MSA: 'inside' is \TrAr{داخل} and 'outside' is \TrAr{خارج}. Table~\ref{table:Leb_sentence} illustrates more examples of Levantine lemmas that are not in MSA, such as \TrAr{عَم}, \TrAr{رَح}, \TrAr{شاف}, \TrAr{لُوما}. 

\subsubsection*{Gloss}
The gloss (\Ar{المعنى بالانجليزية}) is the meaning of the lemma in English.  We restrict the glosses to be SAMA glosses if a SAMA lemma is used, or to Curras if available, otherwise we provide it in the same way.  

\subsubsection*{Stem}
The stem (\Ar{الساق}) is the base word after removing suffixes and prefixes from the token. We follow the $\langle$Stem/POS $\rangle$ tagging schema used in Curras and SAMA, where the stem and the POS are separated by '/'.  The POS is limited to the exact stem POS tagset found in SAMA.

\subsubsection*{Prefixes and Suffixes}
\label{sec:affixes_guidelines}
We follow the prefixes (\Ar{السوابق}) and suffixes (\Ar{االواحق}) tagging schema used in Curras and SAMA: $\{\langle$Prefix1/POS$\rangle+\langle$Prefix2/POS$\rangle \ldots \}$ and $\{\langle$Suffix1/POS$\rangle+\langle$Suffix2/POS$\rangle \ldots \}$. As shown in Table~\ref{table:Leb_sentence} the prefix \Ar{بـ} in the word \TrAr{بروحي} is the preposition \Ar{بـ}/PREP. Multiple prefixes are combined with ``+''. For example, the three prefixes in the word \Ar{وبالقلب} are: \Ar{و}/CONJ+\Ar{بـ}/PREP+\Ar{الـ}/DET.  Suffixes are written in the same way.  For example, the suffixes in the word \TrAr{قلتلهن}   are: \Ar{تـ}/PVSUFF\_SUBJ:1S+ \Ar{لـ}/PREP+\Ar{هن}/PRON\_3FP.  Prefixes and suffixes are critical when dealing with dialects. This is because the morphological difference between dialects and MSA words is mostly due to different combinations of prefixes and suffixes. Dialects use additional types of prefixes and suffixes that are not used in MSA. The prefix \Ar{بـ} in  \Ar{بيضلن}, prefix \Ar{هـ} in \Ar{هالعيون}, or the suffix \Ar{ش} in \Ar{بعرفش}, are examples of affixes that are commonly used in Levantine dialects but are not part of the MSA morphology. To control the quality of our annotations of affixes (i.e., prefixes and suffixes), we extracted the set of all combinations of affixes in Curras and verified them manually (See section ~\ref{sec:Curras_Revisions}). This set along with the SAMA combinations of affixes were then uploaded to the AnnoSheet and used to limit the choices of the annotators.

Table \ref{table:revised_curras_prefixes} presents the set of the prefixes used in the revised version of Curras and uploaded into AnnoSheet to be used by the annotators. Prefixes in Palestinian and Lebanese are all in common but there are two exceptions. The \Ar{ا} in Palestinian can be used  as an INTERROG\_PART, like in \TrAr{اغنيها}, which means ``shall I sing it?''.  However, in such cases in Lebanese, the verb is conjugated in its imperfective first person form to express the same meaning by using \Ar{با} and is said \TrAr{باغنيها}. Additionally, all prepositions in Lebanese and Palestinian are the same, except for \Ar{فـ} which is used only in Palestinian. We would also like to note that there are two prefixes in the corpora that are used only in MSA forms, and not in Palestinian or Lebanese, which are the \Ar{سـ}/FUT\_PART~ and~ the \Ar{لـ}/JUS\_PART. They occur in both corpora due to code-switching as this is a common phenomenon in dialects.

Table~\ref{table:revised_curras_suffixes} presents the set of suffixes used in the revised version of Curras. The majority of of the suffixes are common to both dialects. However, there seems to be one bold systematic difference between the two dialects and it concerns suffixes used to indicate a plural in the 2nd and 3rd person; \TrArSm{هُن}/\TrArSm{كُن} is used in Lebanese to always express a gender-neutral  plural for the 2nd and 3rd person  (e.g., \TrAr{بيتهُن}/\TrAr{بيتكُن}) whereas its Palestinian counterpart uses  \Ar{هِن}/\Ar{كِن} to mostly express a feminine 2nd and 3rd person plural (e.g., \TrAr{بيتهِن}/\TrAr{بيتكِن})  aligning itself with MSA’s \Ar{بيتهُنّ}~\TrNoAr{بيتهُنن}/
\Ar{بيتكُنّ}~\TrNoAr{بيتكُنن}. Nevertheless, the northern Palestinian variety is closer to that of Lebanese and uses \TrArSm{هِن}/\TrArSm{كِن} while remaining gender-neutral. Furthermore, Palestinian uses \TrArSm{كو} where Lebanese uses \TrArSm{كُن} for the 2nd person plural, respectively:   \TrArSm{بيتكو} and \TrArSm{بيتكُن}. Palestinian also tends to use \TrArSm{هم} and \TrArSm{كم},  aligning itself with MSA’s \TrArSm{بيتهم} and \TrArSm{بيتكم}, where Lebanese does not. These occurrences seem to be systematic and may be due to the fact that Lebanese is a Northern Levantine variety while Palestinian is Southern Levantine and such differences are bound to exist in the dialectal continuum, sometimes overlapping in border regions. 

\begin{table}[tb] 
\centering 
\includegraphics[width=0.5\textwidth]{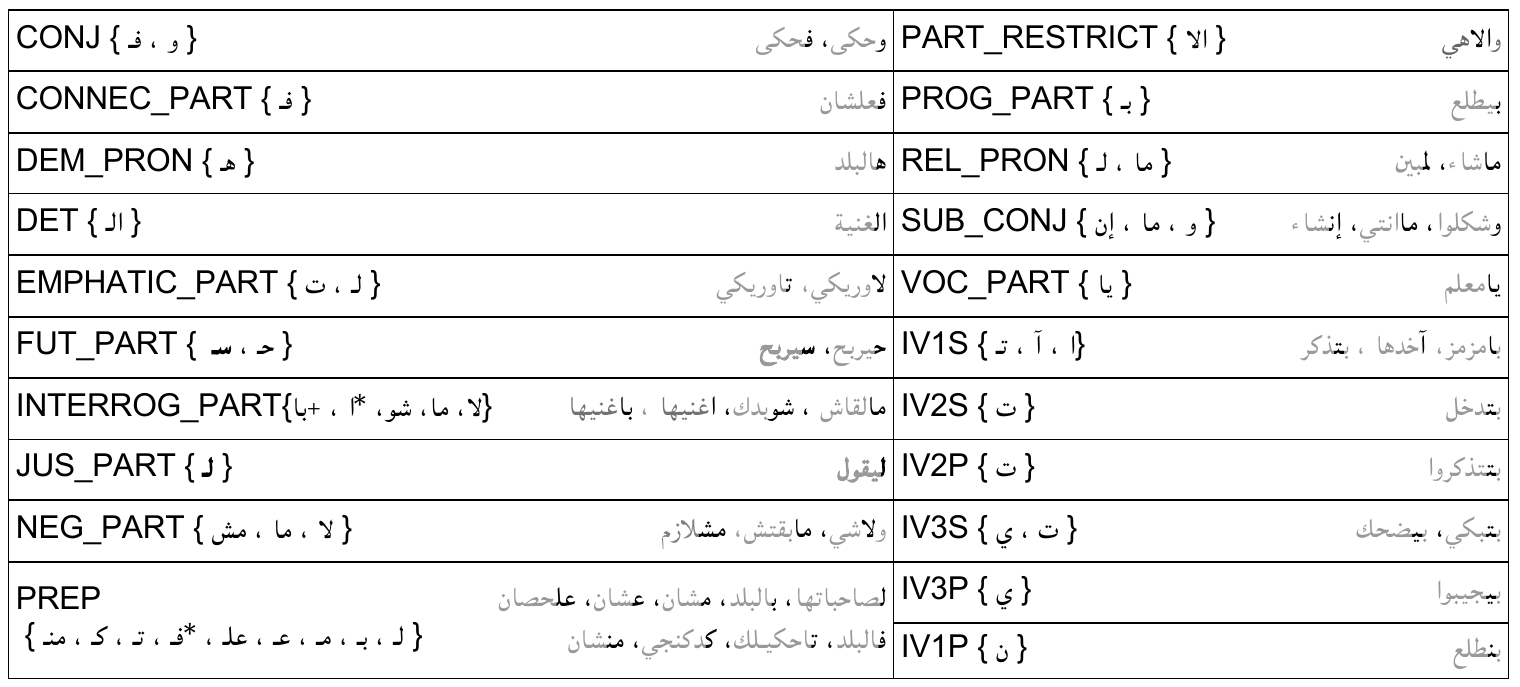}
\caption{List of prefixes and their POS tags in MSA, Palestinian, and Lebanese, which we are used in both corpora. Palestinian-specific prefixes are marked with (*), and Lebanese with (+).
}
\label{table:revised_curras_prefixes}
\end{table} 

\begin{table}[tb]
    \centering
    \includegraphics[width=0.5\textwidth]{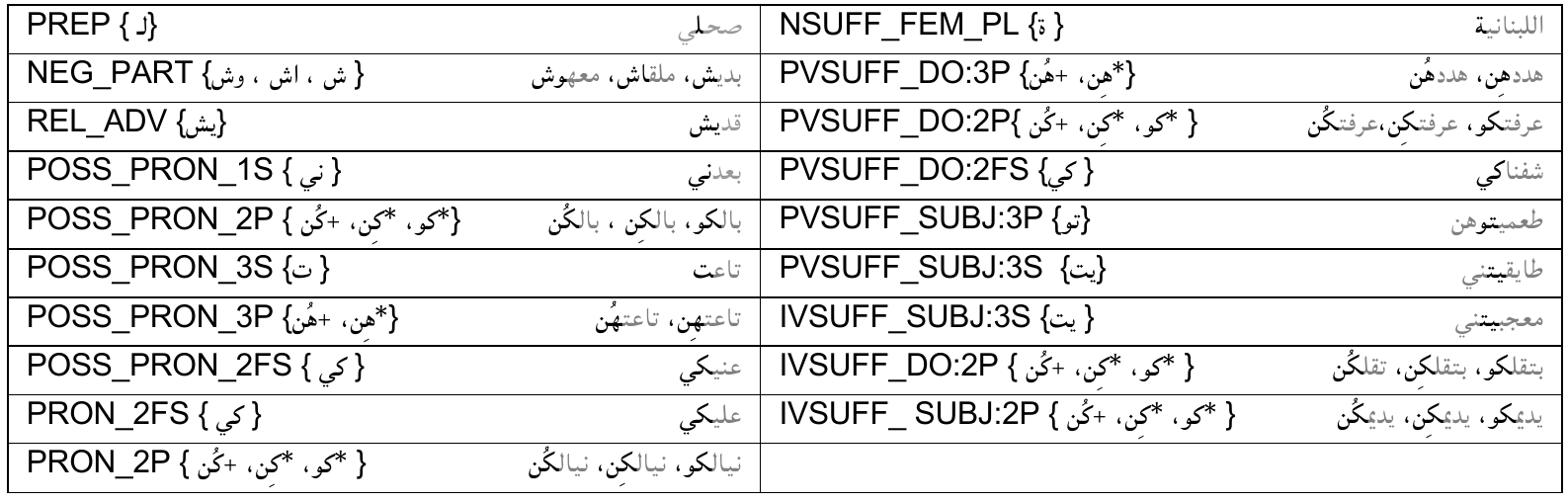}
    \caption{List of suffixes and their POS tags in Palestinian and Lebanese corpora (MSA are excluded as they are many). Palestinian-specific suffixes are marked with (*), and Lebanese with (+).}
  \label{table:revised_curras_suffixes}
\end{table}

\subsubsection*{Part of Speech} 
The part of speech (POS) (\Ar{قسم الكلام}) concerns the grammatical category of the token. The annotators were limited to selecting the POS from the tagset used in SAMA.

\subsubsection*{Person}
The person (\Ar{الإسناد}) refers to the person of the annotated word, if applicable. This can be the $1^{st}$, $2^{nd}$ or $3^{rd}$ person, which we represent by the numbers 1, 2 and 3.

\subsubsection*{Aspect}
This column concerns, for verbs, their aspect (\Ar{صيغة الفعل}). We denote (i) for imperfective verbs (present tense, not completed), (p) for perfective verbs (completed, past tense), and (c) for imperative or command verbs.

\subsubsection*{Gender}
The gender (\Ar{الجنس}) of the word, if applicable, which is (m) for masculine, (f) for feminine and (n) for neutral.

\subsubsection*{Number}
The number (\Ar{العدد}) of the word, if applicable. which is (s) for singular, (p) for plural, or (d) for dual (to count two units).

\section{Evaluation}
\label{sec:eval}

In this section we evaluate the quality of the our annotations for the Lebanese corpus. We performed two evaluations: (\textit{i}) Inter-annotator agreement using the Cohen's Kappa $\kappa$, and (\textit{ii}) The F1-score between each annotator and an expert annotator. The results of the two evaluations are summarized in Tables ~\ref{t:inter_annotation1} and ~\ref{t:inter_annotation2}. To conduct the inter-annotator agreement, we randomly selected annotated sentences that together consist of 400 tokens, i.e., 4.2\% of the corpus. We divided these sentences among our four annotators, such that each annotator re-annotates about 100 tokens that were annotated by another.
We used these 400 new annotations to compute the Cohen's kappa $\kappa$ agreement coefficient. 
The inter-annotation agreement per annotation feature was computed. Table~\ref{t:inter_annotation1} lists the name of the feature and 
the $\kappa$ metric \cite{Eugenio04}:
\[\kappa =\frac{p_o-p_e}{1-p_e} \]

where $p_o$ is the relative observed agreement among annotators and $p_e$ is the hypothetical expected agreement.

In the second evaluation, an expert went over the 400  tokens and corrected the original annotations, if needed. We used these corrections to compute precision and recall where the main expert annotator was considered reference.
The expert annotator performed the following correction actions per feature value:
\begin{itemize}
    \item Approved a feature value annotation (increments $tp:$ {\em true positives} for the feature value)
    \item Approved a missing feature value annotation (increments $tn:$ {\em true negatives} for the feature value)
    \item Rejected a feature value annotation (increments $fp:$ {\em false positives} for the feature value)
    \item Rejected a missing feature value annotation (increments $fn:$ {\em false negatives} for the feature value)
\end{itemize}

The precision $\frac{tp}{tp+fp}$ reflects the ratio of the true positives over the sum of true positives and false positives. The recall $\frac{tp}{tp+fn}$ reflects the ratio of the true positives over the sum of true positives and false negatives. Table \ref{t:inter_annotation2} reports the average precision and recall across all feature values for each feature. We also computed the F1-score based on the precision and recall as:
\[ F1-score = \frac{2 * precision * recall}{precision + recall}\]

The overall kappa $k$, precision, recall, F1-score for all features are calculated using the Python {\tt sklearn.metrics} package. We present the average weighted by the support of each label for precision and recall. 
According to the interpretation of the $\kappa$ score~\cite{j:kappa:McHugh2015}, the aspect and the suffix features scored moderate agreement (between .4 and .6), the stem and the prefix features scored near perfect agreement (above .81), and the rest of the features scored substantial agreement (between .61 and .80). The precision and recall scores of the corrected items show values that concur with the $\kappa$ coefficient. 

\begin{table}
\centering
\small
\begin{tabular}{|wl{1cm}|wr{0.7cm}|wr{1.3cm}|wr{1.6cm}|wr{1cm}|}
\hline 
\thead{Tag}  & \thead{Values}  & \thead{Agreement}  & \thead{Disagreement}  & \thead{Kappa}  \\ \hline 
Stem & 178 & 357 & 43 & 0.884 \\ \hline 
Prefix & 41 & 380 & 20 & 0.860 \\ \hline 
Suffixes & 55 & 358 & 42 & 0.738 \\ \hline 
POS & 22 & 340 & 60 & 0.821 \\ \hline 
Person & 3 & 359 & 41 & 0.629 \\ \hline 
Aspect & 4 & 384 & 16 & 0.911 \\ \hline 
Gender & 3 & 337 & 63 & 0.687 \\ \hline 
Number & 4 & 347 & 53 & 0.741 \\ \hline   \hline 
\thead{Overall} &  &  &  & \thead{0.785} \\ \hline 
\end{tabular}
\caption{Cohen's Kappa coefficient for the inter-annotation agreement and the precision and recall metrics for the main expert corrections.}
\label{t:inter_annotation1}
\end{table}

\begin{table}
\centering
\small
\begin{tabular}{|l|c|c|c|}
\hline 
\thead{Feature}  & \thead{ Precision } & \thead{ Recall } & \thead{ F1-Score }\\ \hline \hline
Stem  & 0.9036 & 0.8935 & 0.893\\ \hline 
Prefixes  & 0.964 & 0.95 & 0.955 \\ \hline 
Suffixes  & 0.948 & 0.895 & 0.915\\ \hline 
POS  & 0.898 & 0.85 & 0.853 \\ \hline 
Person  & 0.928 & 0.898 & 0.910\\ \hline 
Aspect  & 0.974 & 0.96 & 0.967 \\ \hline 
Gender  & 0.845 & 0.843 & 0.844\\ \hline 
Number  & 0.881 & 0.868 & 0.873\\ \hline \hline 
\thead{Overall}  & 0.918 & 0.894 & 0.901\\ \hline 
\end{tabular}
\caption{The precision and recall metrics for the main expert corrections.}
\label{t:inter_annotation2}
\end{table}

These results reflect some areas of disagreement between the annotators. A notable example of this is with prepositions that have a pronoun attached to them such as \Ar{معها} where there should not be any gender or number assigned.
In such an example, some annotators assigned the  gender and number of the suffix \Ar{ها} to token \Ar{مع}. 
That was corrected to be a gender-less and numberless preposition. 
Other disagreements are present in some instances where the suffix \Ar{ة} that indicate the feminine gender are not annotated as a suffix but are merged with the stem of the word. 
Some differences in POS agreement are present for example in the case where gender-less and numberless adverbs are annotated as prepositions or interrogative adverbs (such as \Ar{كيف}) which is not a striking disagreement in itself.

\section{Curras Revisions}
\label{sec:Curras_Revisions}
In order to ensure compatibility with Curras annotations, tagsets, and lemmas, some revisions on Curras were necessary. 
Curras consists of 55,889 tokens. Each token was fully annotated with the morphological features that we adopted in section 5. Since the same token can be used in the same way (i.e., the same features) in different sentences, it is expected that the exact annotations will be repeated. However, we found that this is not always the case in Curras. For example, the same word \TrAr{يدوروا} appeared in two different sentences in Curras, with the same meaning “searching for”, but each time with a different MSA lemma: (\Ar{بَحَث}) and  (\Ar{بحث}); while it should be (1 \Ar{بَحَثَ۪}). The adverb \TrAr{بس} was
correctly annotated in all occurrences in Curras; however, in some cases, it was mistakenly assigned with gender; and in some cases, it was annotated with the noun POS. We also found some typos in the tagsets of the stems and affixes.

Our goal is to unify and normalize such variations, and then build a list of morphological solutions as clean as possible.We performed the following revision steps:\\

\textbf{a. Tokenization and POS} \\
We developed a POS parser that reads the prefixes, stem, and suffixes in a given solution (i.e., annotations of a token), and returns a validation flag. We carefully inspected solutions that were flagged for review. 
The POS parser validates the following: (\textit{i}) no parsing errors in the prefixes, stem, and suffixes, (\textit{ii}) the transliterations of the prefixes, stem, and suffixes in Buckwalter are correct, (\textit{iii}) the concatenation of the prefixes, stem, and suffixes corresponds to that of the CODA, (\textit{iv}) every prefix should be in the predefined set of prefixes, (\textit{v}) every suffix should be in the predefined set of suffixes, and (\textit{vi}) every stem POS should be in the SAMA POS tagset.\\

\textbf{b. Lemmatization}

Curras originally contained 8,560 unique lemmas. 
Although Curras was annotated using SAMA lemmas, some of Curras lemmas were incorrectly linked with SAMA lemmas. 
This was mostly because of partial diacritization of lemmas (e.g., \Ar {بحث}) or as the lemma subscript is ignored (e.g., \Ar {بَحَث}).
Ignoring diacritics and subscripts makes the lemma ambiguous. Thus, we cannot know, for example, whether it is (1 \Ar {بَحَثَ۪}), (2 \Ar {بَحْث}), or (1 \Ar {بَحْث}). To disambiguate MSA lemmas in Curras and link them with SAMA lemmas, we developed a lemma disambiguator that takes the lemma, POS, and gloss, and tries to reduce the number of choices. 
In case one lemma is returned, it is then considered the correct SAMA lemma, otherwise undecided. We were able to disambiguate about 5,120 unique lemmas (i.e., 58\%) in this way. The remaining undecided 3,560 lemmas were manually disambiguated and linked with SAMA. As a result, the unique number of MSA lemmas in Curras now is 7,313. These include 6,781 that are mapped with SAMA lemmas, and 432  MSA lemmas that are not found in SAMA. We marked the latter with ``\_0''.

Validating and unifying dialect lemmas was straightforward. In case a dialect lemma has the same letters as the MSA lemma (i.e., ignoring diacritics and subscripts) then it is the same lemma. So, we replace the dialect lemmas with the MSA lemma. Otherwise, a manual verification is performed. As a result, the unique number of dialect lemmas in Curras is 8,510. These include 7,785 lemmas equivalent to MSA lemmas, and 1,012 dialect lemmas that have no corresponding MSA lemmas. We also marked the latter with ``\_0''.\\

\textbf{c. Other features}

We applied some heuristics in cleaning the Person, Aspect, Gender, and Number features.
For example, the Aspect and Person are assigned only to verbs, otherwise they should be ``-''. We compared the Gender and Number with the suffix tags which also indicate gender and number, and corrected mistakes manually when needed.\\

\textbf{d. Generating Unique Solutions}

We prepared a table with unique annotations from Curras, called the ``Solutions'' table. 
We reused these solutions to annotate the Lebanese corpus in order to maximize the compatibility between both corpora. 
To do this, we split Curras into two tables: Tokens and Solutions, with a solution identifier (id) to link them. 
The Tokens table contains only the token id, token, and solution id. 
In this manner, the tokens that have the exact same annotations are given the same solution id. 
The Solutions table contains all annotations after removing the exact redundancies, which consists of 16,244 solutions. 
We considered two solutions to be identical if they have the same: CODA, prefixes, stem, suffixes, DA lemma, MSA lemma, Person, Aspect, Gender, and Number.
We uploaded the Solutions table into our AnnoSheet and enabled our annotators to look up and re-use annotations from the Solutions table, as described in Section~\ref{sec:annotation_methodology}. 
As a result of this effort, we envision that the revised version of Curras along with the additions from Baladi; the newly built Lebanese corpus, 
form a more Levantine dialect corpus.

\section{Discussion: a more Levantine Corpus}
\label{sec:discussion}
In this section, we discuss how both Palestinian and Lebanese corpora can be used as one, more Levantine corpus. Not only they are annotated with the same tagsets as discussed earlier, but adding 9.6K annotated Lebanese tokens to the Palestinian corpus Curras has helped bridge the nuanced linguistic gaps that exist between the two highly mutually intelligible dialects. Those nuances, as discussed earlier in this paper, are notably present in the affixes (i.e., morphology). Indeed, some prefixes and suffixes are typically Palestinian and not habitually used in Lebanese and vice-versa. However, these differences are a few and the majority of affixes are common to both dialects (See the differences in section \ref{sec:affixes_guidelines}).

Additionally, Lebanese functional words have also been incorporated, solidifying our idea of a more Levantine corpus where the dialectal continuum is taken into account. In fact, the majority of the functional words are common to both dialects. Table \ref{fig:leb_pal} presents frequent functional words that are different in both dialects and the mapping between them. 
To summarize, both corpora consists of about 65.2K tokens, covering both Palestinian and Lebanese, annotated using the same guidelines.

\begin{table}[h]
    \centering
    \includegraphics [scale=0.9]{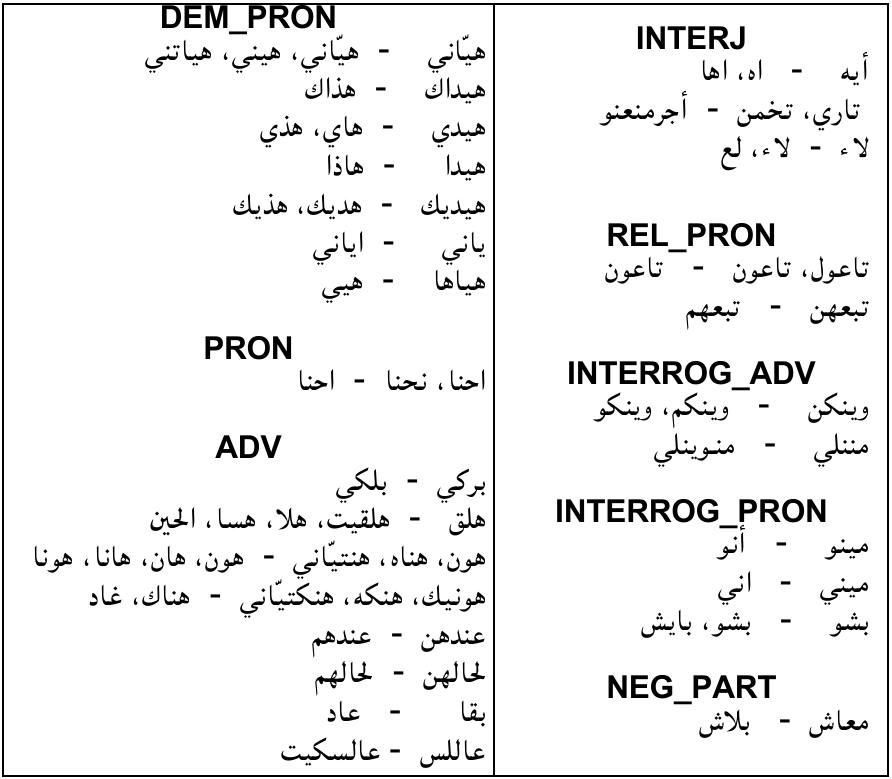}
    \caption{Frequent functional words in Lebanese (right) and Palestinian (left).}
    \label{fig:leb_pal}
\end{table}

\section{Conclusions and Future Work}
\label{sec:conclusion}

In this paper we presented the first morphologically annotated corpus for the Lebanese dialect 9.6K tokens. We also present a revised version of Curras, the Palestinian dialect corpus, about 55.9K tokens. We also described the various challenges we faced and measures we took to produce a compatible and more general Levantine Corpus, consisting of 55.9K tokens annotated with rich morphological and semantic information. Still, the evaluation of our annotators’ performance shows a high degree of consistency and agreement. The Lebanese corpus is available for downloading and browsing online.

We plan to increase the size of our corpus to cover additional Levantine sub-dialects, especially those of other Levantine areas, most notably some of Syria’s dialectal varieties. We also plan to use this corpus to develop morphological analyzers and word-sense disambiguation system for Levantine Arabic as we did for MSA (see \cite{HJ21b,HJ21}). Additionally, we plan to build on the Palestinian and Lebanese dialect lemmas to develop a Levantine-MSA-English Lexicon and extend it with synonyms \cite{JKKS21}. Both Curras and Baladi corpora are also being annotated with named-entities as part of the Wojood NER corpus see \cite{JKG22}.

\section{Acknowledgements}
This research is partially funded by the Palestinian Higher Council for Innovation and Excellence. The authors also acknowledge the great efforts of many students who helped in the annotation process, especially Tamara Qaimari, Shimaa Hamayel, Dua Shwiki, Asala Hamed, and Ahd Nazeeh.

\section{Bibliographical References}
\label{reference}

\bibliographystyle{lrec2022-bib}
\bibliography{references, MyReferences}
\bibliographystylelanguageresource{lrec2022-bib}
\end{document}